\documentclass[letterpaper, 10 pt, conference]{ieeeconf}  
 \pdfoutput=1

\IEEEoverridecommandlockouts                              

\usepackage{stfloats}
\usepackage{subcaption}
\usepackage{mathtools}
\usepackage{graphics} 
\usepackage{epsfig} 
\usepackage{mathptmx} 
\usepackage{times} 
\usepackage{amsmath} 
\usepackage{amssymb}  
\usepackage{multicol}
\usepackage{graphicx}
\usepackage{epstopdf}
\usepackage{pdfpages}
\usepackage{cite}
\usepackage{amsbsy}
\usepackage{bm}
\usepackage{accents}
\newcommand*{\dt}[1]{%
  \accentset{\mbox{\bfseries .}}{#1}}

\usepackage{cite}
\usepackage{booktabs}
\pdfoutput=1
\usepackage{hyperref}
\hypersetup{
    colorlinks=true,
    linkcolor=black,
    citecolor=black,
    filecolor=black,
    urlcolor=black,
}

\def\horizontaldistance{\kern2pt}

\makeatletter
\newcommand*{\rom}[1]{\expandafter\@slowromancap\romannumeral #1@}
\makeatother
\title{\LARGE \bf
Push Recovery of a Position-Controlled Humanoid Robot Based on Capture Point Feedback Control
}

\author{Milad Shafiee-Ashtiani,$^{1}$ Aghil Yousefi-Koma,$^{1}$ Reihaneh Mirjalili,$^{1}$ Hessam Maleki and Mojtaba Karimi$^{1}$ 
\thanks{$^{1}$Center of Advanced Systems and Technologies (CAST)
School of Mechanical Engineering, College of Engineering, University of Tehran, Tehran, Iran.
       ( {\tt\small shafiee.a@ut.ac.ir}) }%
}

\begin{document}

\maketitle
\thispagestyle{empty}
\pagestyle{empty}

\begin{abstract}
In this paper,  a combination of ankle and hip strategy is used  for push recovery of a position-controlled humanoid robot. Ankle strategy and hip strategy are equivalent to Center of Pressure (CoP)  and Centroidal Moment Pivot (CMP) regulation respectively. For controlling the CMP and CoP we need a torque-controlled robot, however most of the conventional humanoid robots are position controlled. In this regard, we present an  efficient way for implementation of the hip and ankle strategies on a position controlled humanoid robot. We employ a feedback controller to compensate the capture point error. Using our scheme, a simple and practical push recovery controller is designed which can be implemented on the most of the conventional humanoid robots without the need for torque sensors. The effectiveness of the proposed approach is verified through push recovery experiments on SURENA-Mini humanoid robot under severe pushes.
\end{abstract}

\section{INTRODUCTION}
The main target of legged robots research is realizing a  robot that is able to work in real environments. Real  environments  are cluttered and in these situations different disturbances will applied  on the robot body. Biped robots have unstable nature because of unilaterally constraint beetween  foot and ground. Thus, for a biped robot the ability of recovering from unexpected external disturbances is essential.

In recent years, several attempts have been made by researchers of the field to generate robust motion of biped robots (\cite{kajita2003biped,wieber2006trajectory,pratt2006capture,herdt2010online,stephens2010push,aftab2012ankle,koolen2012capturability}). A common way for ensuring dynamic balance during walking is to maintain the Zero Moment Point (ZMP) or the Center of Pressure (CoP) within the support polygon of the feet and ground. 
The conventional approaches that have been used for robust motion control of humanoid robots are based on the Model Predictive Control (MPC) or controlling the Capture Point (CP) \cite{kajita2003biped,wieber2006trajectory,pratt2006capture,herdt2010online,stephens2010push,aftab2012ankle,koolen2012capturability,englsberger2015three,krause2012stabilization,yun2011momentum}.

Kajita et al.\cite{kajita2003biped} presented an efficient preview controller  as a first successful attempt for MPC based  motion generation methods. This method was expressed more generally as an MPC problem by Wieber et al. \cite{wieber2006trajectory}. In order to increase the robustness of the controller, the MPC formulation in \cite{wieber2006trajectory} has been modified to adapt the step position automatically \cite{herdt2010online,stephens2010push} and step timing \cite{khadiv2016step}. In this regard, Shafiee et. al \cite{shafiee2017robust} proposed a MPC scheme that combines all the ankle, hip, and stepping strategies for balance recovery of humanoid robots in a unit formulation.

Pratt et al. \cite{pratt2006capture,koolen2012capturability} introduced the CP as a unstable part of the CoM dynamics. Therefore, it requires a controller for stabilizing unstable nature of dynamic of the CP. To this end, Englsberger et al. \cite{englsberger2015three,krause2012stabilization} developed a controller for CP tracking.
\begin{figure}
 \includegraphics[,scale=0.54, trim ={2.9cm 16.cm 6.3cm 2cm},clip]{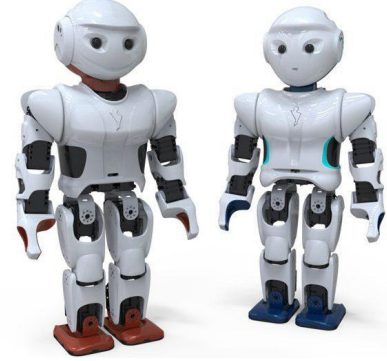}\par 
\caption{SURENA-Mini a position-controlled  humanoid robot is developed at CAST}
 \label{fig111}
\end{figure}

The centroidal angular momentum plays a vital role for balance recovery\cite{herr2008angular} especially in the situation that stepping is not possible or contact surface is limited \cite{yun2011momentum,kiemel2012balance,shafiee2016push}.  
Accordingly, we need to modulate both of the CMP and CoP for  maintaining the balance of a biped robot. 

Controlling the CMP or CoP in torque-controlled humanoid robots is usual, however, exact control of the  CoP on position-controlled humanoid is very difficult \cite{del2016implementing} and unfortunately most of the conventional humanoid robots such as SURENA-Mini humanoid robot are position-controlled(Fig.\ref{fig111}). Therefore implementation of the push recovery based on the mentioned MPC scheme on a position controlled humanoid robot is challenging.

In this paper, in order to implement a push recovery controller on a position-controlled humanoid robot, the CP concept is used in a position-based feedback controller. To do so, a simple and efficient feedback controller is developed for push recovery without the need of torque sensor at each joint or force/torque sensors at feet. The main goal of this controller is to compensate the CP error by using the ankle and hip strategy. The experimental results shows the controller can save the robot from falling under severe pushes. 

The remainder of this paper is organized as follows. The CoM dynamics, and the CP formulations are reviewed in Sec  \rom{2}. The proposed push recovery controller is presented in Sec  \rom{3}. In  \rom{4}, the obtained experiments results are presented and discussed. Finally, Section  \rom{5} concludes the findings.
\begin{figure*}
 \includegraphics[,scale=0.8, trim ={0.1cm 21.85cm 1cm 1.8cm},clip]{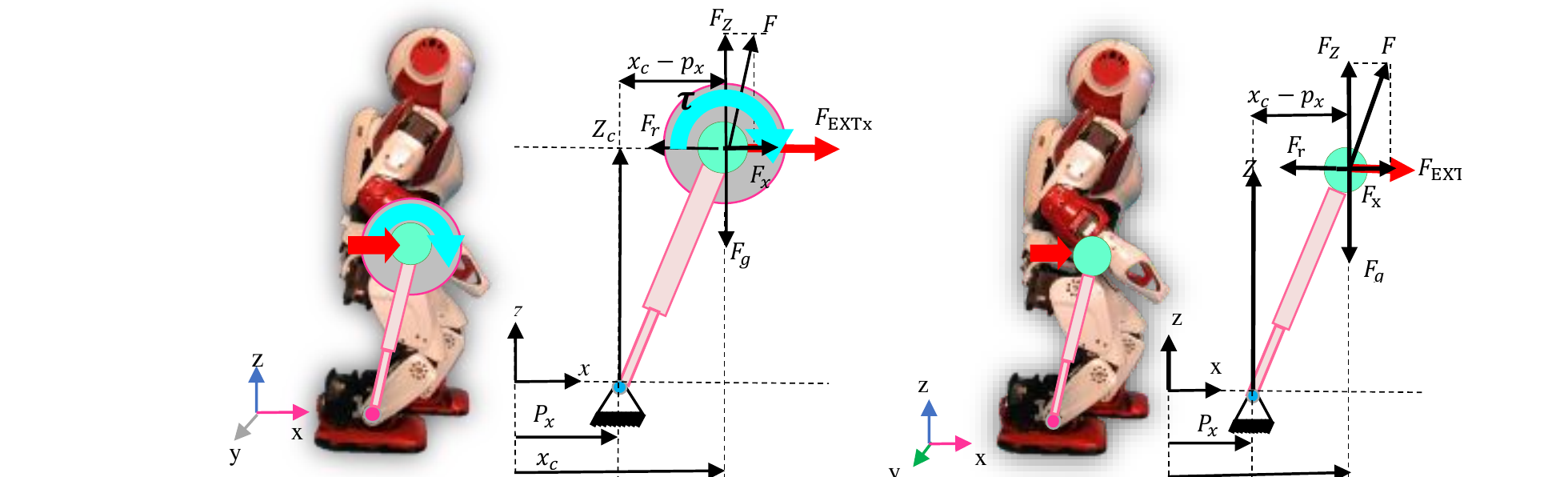}\par 
\subcaptionbox{Linear inverted pendulum+Flywheel model}[29em]{\centering } \subcaptionbox{Linear inverted pendulum model}[14em]{\centering }
  \hfill\null
\caption{Linear Centroidal Dynamics of humanoid robot}
 \label{fig2}
\end{figure*}

\section{Center of Mass Dynamics}

Using the centroidal dynamics of a biped robot for motion planning makes the corresponding optimization problem non-convex \cite{orin2013centroidal}. 
The centoridal dynamics of a biped robot can be approximated by the Linear Inverted Pendulum Model (LIPM) \cite{kajita20013d}. This model is a good dynamic approximation of a biped robot, particularly for the standing posture. The LIPM uses the following assumptions \cite{kajita20013d}:

\begin{itemize}
\item The rate of change of  centroidal angular momentum is zero,
\item The CoM motion constraints to move on a plane with constant height
\end{itemize}

Based on the mentioned assumptions and Fig.\ref{fig2}, the equation of motion of the LIPM will be expressed as follows:

\begin{equation}
\begin{aligned}
\ddot x_{c}= \omega_n^{2}( x_{c}-p_{x}) \hspace{1.1cm}\\
\ddot y_{c}= \omega_n^{2}( y_{c}-p_{y}) \hspace{1.1cm}\\
\end{aligned}
 \label{eq1}
\end{equation}
where $m$ is the robot mass, the  position of CoM and CoP(ZMP) is expressed by $P_c=[x_c,y_c,z_c]^T$ and $ P_{CoP}=[p_x, p_y, 0]^T$ respectively and $\omega_{n}=\sqrt{(g/z_c )}$ is the natural frequency of the LIPM. The Ground Reaction Force (GRF) intersects with the CoM because the base joint of the LIPM is without torque and the derivation of centroidal angular momentum is zero. Based on the Fig.\ref{fig2}, $F_z$ is the vertical component of the GRF. It is equal with the gravitational force $F_g$ that acts on the CoM. The inertial force $ F_r=m\ddot x _c$  completes the equilibrium of forces in $P_c$.

The centroidal angular momentum, especially the generated angular momentum by torso and arms, can play a vital role in balance recovery. These joints can be used to apply a moment about the CoM. In the case of zero torque about the CoM, the CMP and CoP positions will be equal. For a non-zero torque about the COM, however, the CMP can goes beyond the edge of the foot, while the COP still locates within the contact region of feet. This centroidal angular momentum can be approximated by considering the upper body as a flywheel as shown by Pratt \cite{pratt2006capture}. In other words, the CMP is the point where a line parallel to the ground reaction force and passing through the COM intersects the ground. Therefore, by adding this linear approximation of centroidal angular momentum to the LIPM dynamics, the equations of motion of LIPM+flywheel can be expressed as(Fig. \ref{fig2}):

\begin{equation}
\begin{aligned}
\ddot x_{c}= \omega_n^{2}( x_{c}-p_{x}) -\frac{\dot H_{y}}{mz}\hspace{1.1cm}\\
\ddot y_{c}= \omega_n^{2}( y_{c}-p_{y}) +\frac{\dot H_{x}}{mz}\hspace{1.1cm}\\
\end{aligned}
 \label{eq2}
\end{equation}
where $\dot{H}$ is the rate of centroidal angular momentum that can be generated by the torque of upper-body joints(arm and trunk joints). The relation between the CoP and the CMP can be written as: \cite{kajita20013d}:

\begin{equation}
\begin{aligned}
CMP_{x} =p_{x}+\frac{\dot H_{y}}{F_z}\hspace{1.1cm}\\
CMP_{y} =p_{y}-\frac{\dot H_{x}}{F_z}\hspace{1.1cm}\\
\end{aligned}
 \label{eq3}
\end{equation}

In this regard, by combining (\ref{eq2}) and (\ref{eq3}), we will have:
\begin{equation}
\begin{aligned}
\ddot x_{c}= \omega_n^{2}( x_{c}-CMP_{x}) \hspace{1.1cm}\\
\ddot y_{c}= \omega_n^{2}( y_{c}-CMP_{y}) \hspace{1.1cm}\\
\end{aligned}
 \label{eq4}
\end{equation}

When a severe disturbance  acts on the robot, the CP might go outside of support polygon. Therefore the CMP and CoP(ZMP) will diverge and the CMP can leave the support polygon for controlling the CP by generating centroidal angular momentum. 

\subsection{Relation between CoM, CP and CoP }

The unstable part of the CoM dynamics has been called the CP and is defined as follows \cite{pratt2006capture,koolen2012capturability,englsberger2015three}: 
\begin{equation}
\begin{aligned}
{\xi_x}={x_c}+\frac{{\dt x_c}}{{\omega_n}} \hspace{1.1cm}\\
{\xi_y}={y_c}+\frac{{\dt y_c}}{{\omega_n}} \hspace{1.1cm}\\
\end{aligned}
  \label{eq5}
\end{equation}
From (\ref{eq5}), the CoM dynamics is given by:

\begin{equation}
\begin{aligned}
{\dot x_c}={{\omega_n}} ({\xi}-{ x_c} )\hspace{1.1cm}\\
{\dot y_c}={{\omega_n}} ({\xi}-{ y_c} )\hspace{1.1cm}\\
\end{aligned}
  \label{eq6}
\end{equation}


By differentiating (\ref{eq6}) and substituting (\ref{eq1}) the CP dynamics is given by:

\begin{equation}
\begin{aligned}
\dt{\xi}_x=\omega_n({\xi_x}-{p}_{x})\hspace{1.1cm}\\
\dt{\xi}_y=\omega_n({\xi_y}-{p}_{y})\hspace{1.1cm}\\
\end{aligned}
  \label{eq7}
\end{equation}

Also by considering the effect of centroidal angular momentum based on (\ref{eq4}) the (\ref{eq7}) will be modified as follows:

\begin{equation}
\begin{aligned}
\dt{\xi}_x=\omega_n({\xi_x}-{CMP}_{x})\hspace{1.1cm}\\
\dt{\xi}_y=\omega_n({\xi_y}-{CMP}_{y})\hspace{1.1cm}\\
\end{aligned}
  \label{eq8}
\end{equation}

As it is obvious in (\ref{eq7}), the  CP has a unsatble first order dynamic and CMP always  push the CP. In order to recover the balance of a humanoid robot, the CP should be controlled. When the CP is located within the support polygon, it can be controlled by the CoP \cite{englsberger2015three}, and when it is located out of the support polygon it can be controlled by the CMP or stepping.

Using the concept of CP we can determine when and where to take a step to recover from a push \cite{pratt2006capture}. If the CP is located within the support polygon, the robot is able to recover from the push without having to use hip strategy or take a step. In order to stop in one step, the support polygon must have an intersection with the capture region  \cite{pratt2006capture}. The robot will fail to recover from a severe push in one step, if the capture region does not intersect with the kinematic workspace of the swing foot. In the next sections we will discuss how to use the CP in Push recovery controller based on the ankle and hip strategy.

\section{Human-Inspired Balancing Strategies}

\begin{figure}[]
\centering  
 \includegraphics[scale=0.57, trim ={10.5cm 6.9cm 11cm 5.4cm},clip]{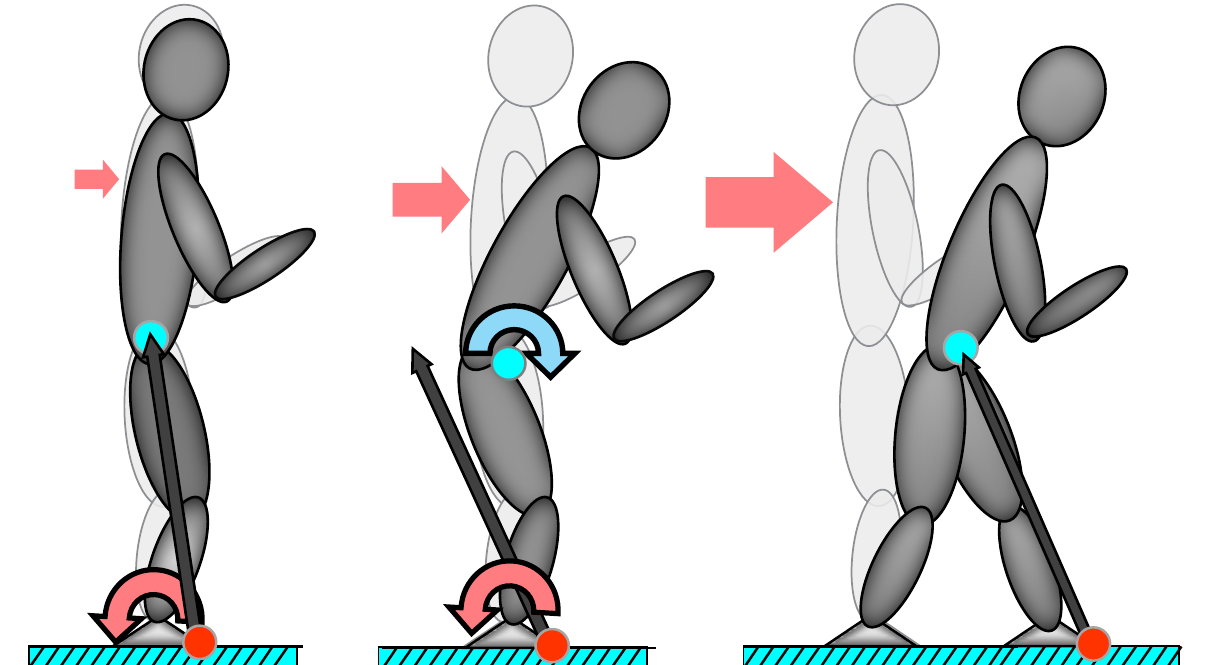}
\hfill\subcaptionbox{Ankle strategy}[7em]{\centering}
 \subcaptionbox{Hip strategy}[8em]{\centering } \subcaptionbox{Stepping strategy}[8em]{\centering }
  \hfill\null
\caption{Human-inspired balancing strategies}
      \label{fig3}
    \end{figure}

The response of a human to progressively increasing disturbances can be categorized into three basic strategy: (1) ankle strategy, (2) hip strategy (3) and stepping strategies. 
Humans tend to use the ankle strategy in case of small pushes to bring back the CP to its desired position as depicted in Fig.\ref{fig3}(a). However, the contact between the foot and floor has a unilateral constraint and if the ankle torque becomes  large, the CoP locates on the edge of the support polygon and the foot starts to rotate. Angular momentum of the upper body can be generated in the direction of the disturbance by applying a torque on the hip joint or arm joint as shown in Fig.\ref{fig3}(b).  This strategy is also called CMP Balancing. With increasing the disturbance the useful strategy will be stepping Fig.\ref{fig3}(c).  However,  several situations might occur where stepping is not possible. In this situation the balance recovery by Hip-Ankle strategy is necessary \cite{kiemel2012balance}. In this paper, the hip and ankle strategy is deployed by a feedback controller. 
\subsection{Ankle strategy}  
Ankle strategy is equivalent to CoP balancing and is used for controlling the CoP in support polygon. In fact most of the time it can be controlled by torque of ankle joint but it is not limited to this because if we assume that ankle joint is locked and we don't have actuation in ankle joint, we can control the CoP with knee or other joints. Therefore the Ankle strategy is not limited to use ankle joint. Briefly it is equivalent to change linear momentum of the centroidal dynamic of a humanoid robot. 

The relation between position of CoP and torque of ankle joint is as follows:
\begin{equation}
\begin{aligned}
p_{x}=\frac{{\tau}_{y,ankle}}{{mg}}\hspace{1.1cm}\\
p_{y}=\frac{{\tau}_{x,ankle}}{{mg}}\hspace{1.1cm}\\
\end{aligned}
  \label{eq9}
\end{equation}
 By substituting the (\ref{eq9}) into (\ref{eq1}) we will have:
 
 \begin{equation}
\begin{aligned}
\ddot x_{c}= \omega_n^{2}( x_{c}-\frac{{\tau}_{y,ankle}}{{mg}}) \hspace{1.1cm}\\
\ddot y_{c}= \omega_n^{2}( y_{c}-\frac{{\tau}_{x,ankle}}{{mg}}) \hspace{1.1cm}\\
\end{aligned}
 \label{eq10}
\end{equation}

let us assume that we have a reference trajectory for capture point, therefore we will have the following equation:
\begin{equation}
\begin{aligned}
\dt{\xi}_{x,ref}=\omega_n({\xi_{x,ref}}-{p}_{{x,ref}})\hspace{1.1cm}\\
\dt{\xi}_{y,ref}=\omega_n({\xi_{y,ref}}-{p}_{y,ref})\hspace{1.1cm}\\
\end{aligned}
  \label{eq11}
\end{equation}

let us define the error of CP as $\dt{\xi}_{error}=\dt{\xi}_{ref}-\dt{\xi}$    , therefore by  subtraction of (\ref{eq11}) from (\ref{eq7}) we will have:

\begin{equation}
\begin{aligned}
\dt{\xi}_{x,error}=\omega_n({\xi_{x,error}}-{p}_{{x,error}})\hspace{1.1cm}\\
\dt{\xi}_{y,error}=\omega_n({\xi_{y,error}}-{p}_{y,error})\hspace{1.1cm}\\
\end{aligned}
  \label{eq12}
\end{equation} 

By substituting (\ref{eq9}) into (\ref{eq12}) we will have:
\begin{equation}
\begin{aligned}
\dt{\xi}_{x,error}=\omega_n({\xi_{x,error}}-\frac{{\tau}_{y,ankle}}{{mg}})\hspace{1.1cm}\\
\dt{\xi}_{y,error}=\omega_n({\xi_{y,error}}-\frac{{\tau}_{x,ankle}}{{mg}})\hspace{1.1cm}\\
\end{aligned}
  \label{eq13}
\end{equation}

Considering the torque of ankle joint as a control input (\ref{eq13}), the error of CP can be compensated  with the following PD controller:
 \begin{equation}
\begin{aligned}
{\tau}_{y,ankle}=k_p{\xi_{x,error}}+k_d{\dt{\xi}_{x,error}}\hspace{1.1cm}\\
{\tau}_{x,ankle}=k_p{\xi_{y,error}}+k_d{\dt{\xi}_{y,error}}\hspace{1.1cm}\\
\end{aligned}
  \label{eq14}
\end{equation} 

\begin{figure*}
\begin{multicols}{2}
    \includegraphics[,scale=0.96, trim ={1.3cm 18.2cm 0.5cm 7.8cm},clip]{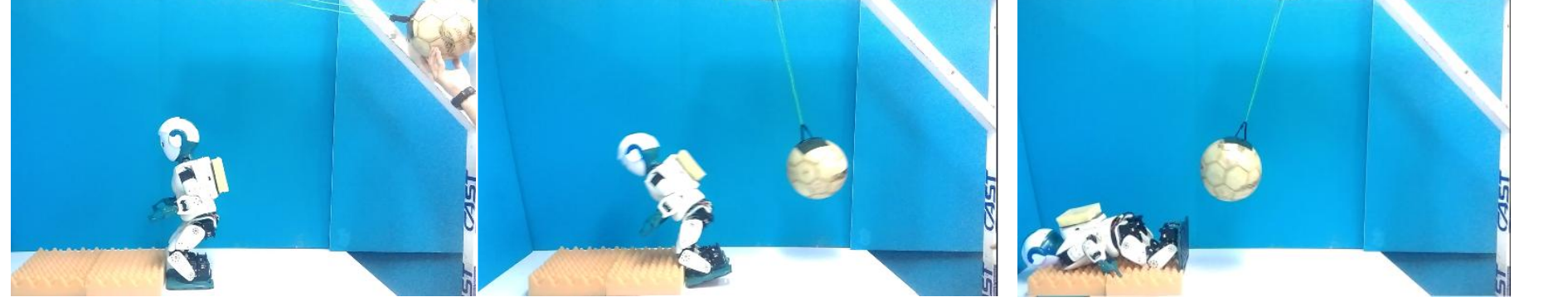}\par
\end{multicols}
\caption{ Push recovery behavior of the robot without controller}
 \label{fig5}
\end{figure*}

\begin{figure*}
\begin{multicols}{2}
 \includegraphics[,scale=0.95, trim ={1.cm 14.5cm 1cm 7.7cm},clip]{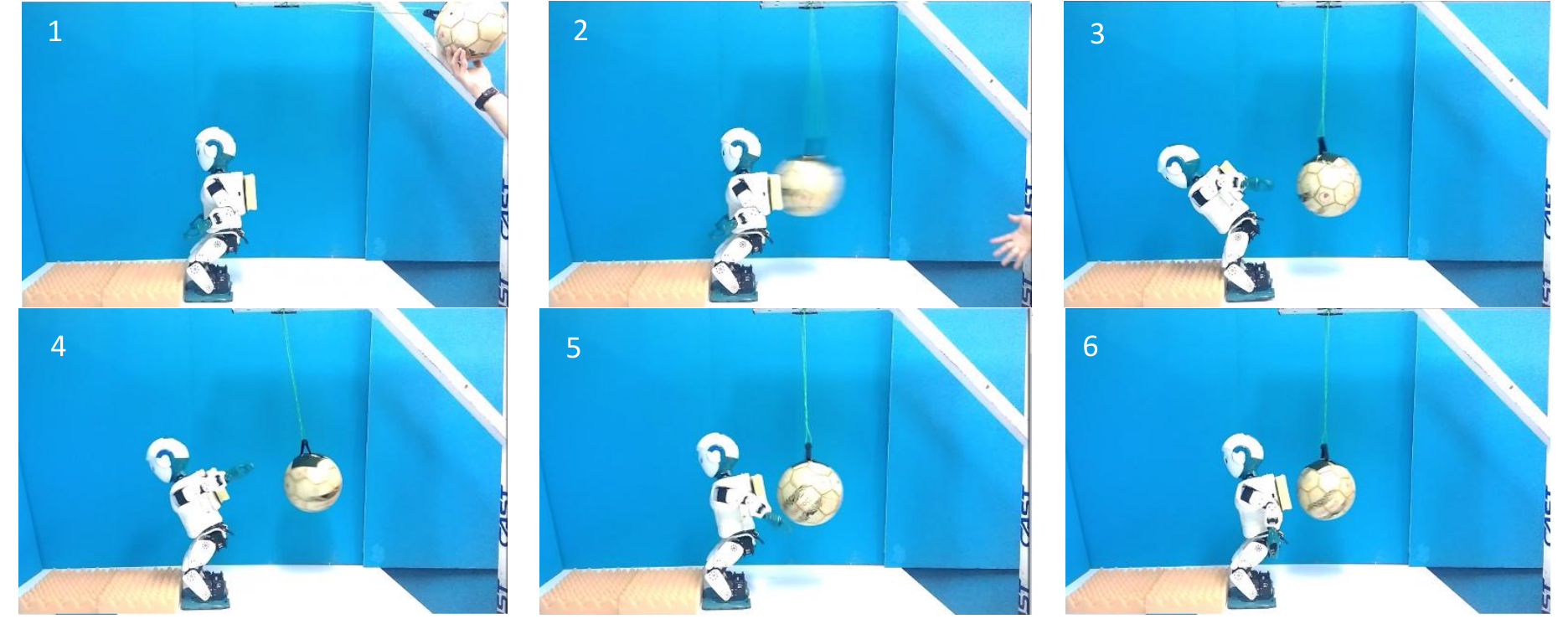}\par 
    \end{multicols}
\caption{Push recovery behaviour of the robot with controller. (A video of this push recovery behavior experiment is publicly available online.) \protect\footnotemark }
 \label{fig4}
\end{figure*}

Therefore CP can be controlled by ankle torque, however there is limitation to the torque of ankle because of unilateral constraint that CoP should be placed inside of support polygon. The greater the CP error, the more ankle torque is needed. This causes CoP to go out of support polygon and foot  will tip over. Therefore If CP goes out of support polygon the robot will need to use hip strategy and generate angular momentum to control the CP to the desired position.

\begin{figure*}
\begin{multicols}{2}
 \includegraphics[,scale=0.87, trim ={6cm 15.8cm 6cm 3.4cm},clip]{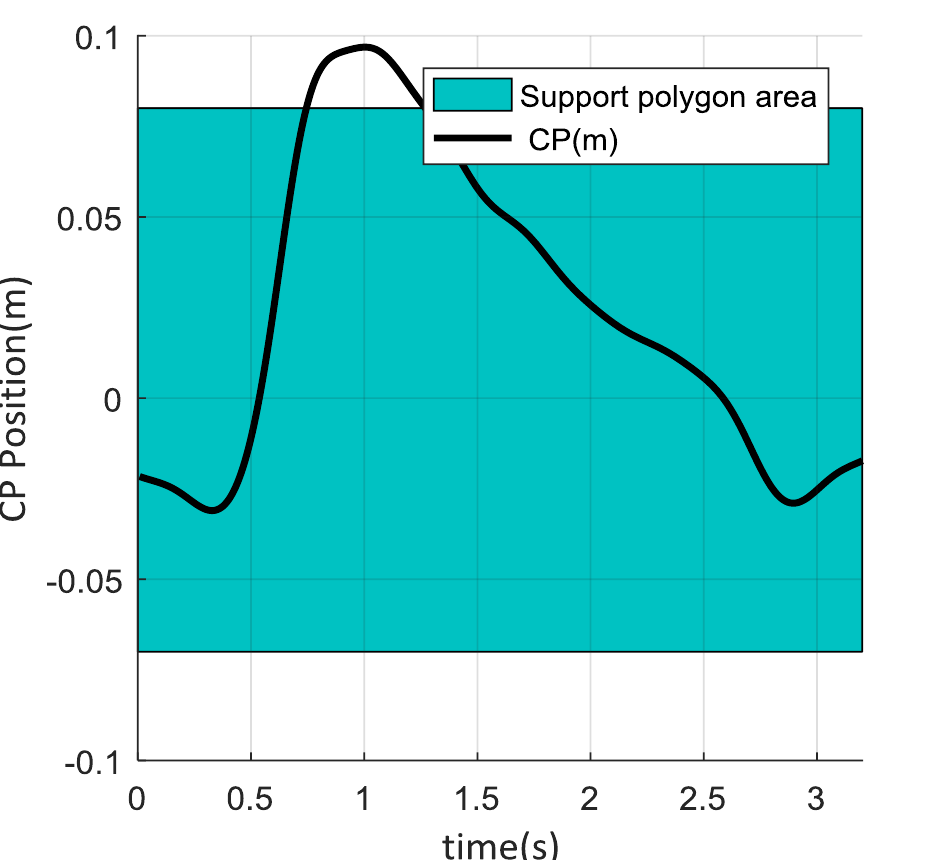}\par 
    \includegraphics[,scale=0.81, trim ={5cm 15.5cm 6cm 3.2cm},clip]{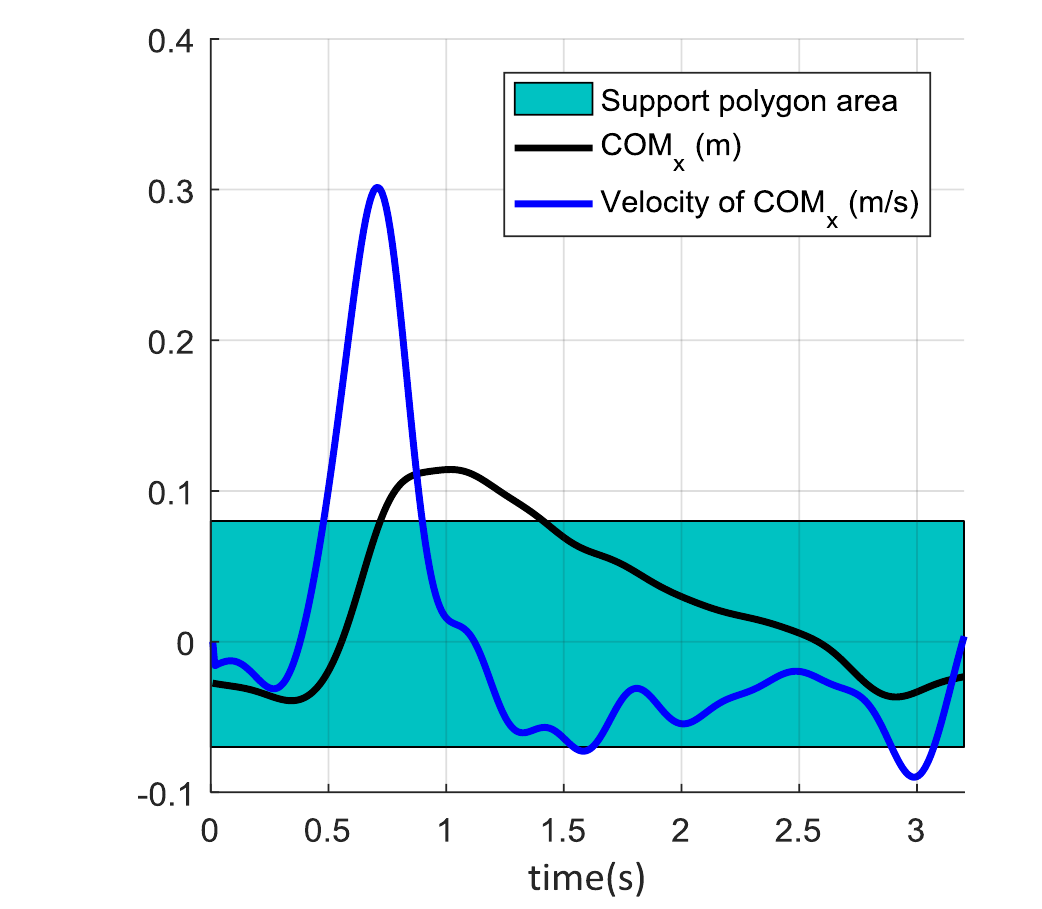}\par 
    \end{multicols}
\begin{multicols}{2}
    \includegraphics[,scale=0.85, trim ={5.8cm 15.8cm 6.8cm 3.8cm},clip]{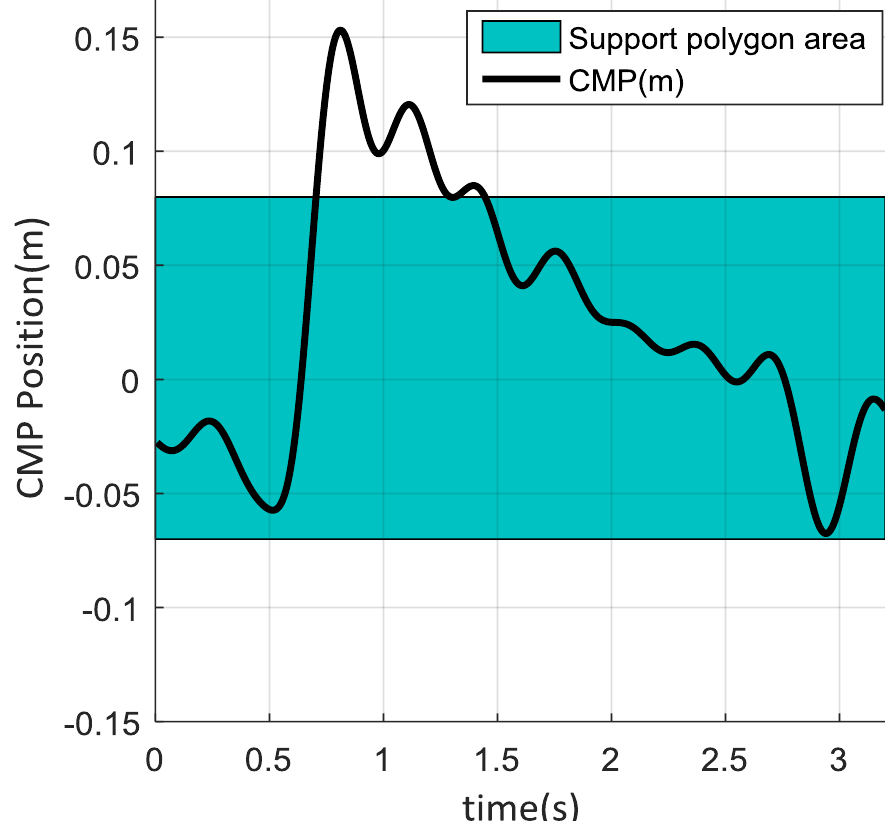}\par
 \includegraphics[,scale=0.85, trim ={5.5cm 15.9cm 6.5cm 3.7cm},clip]{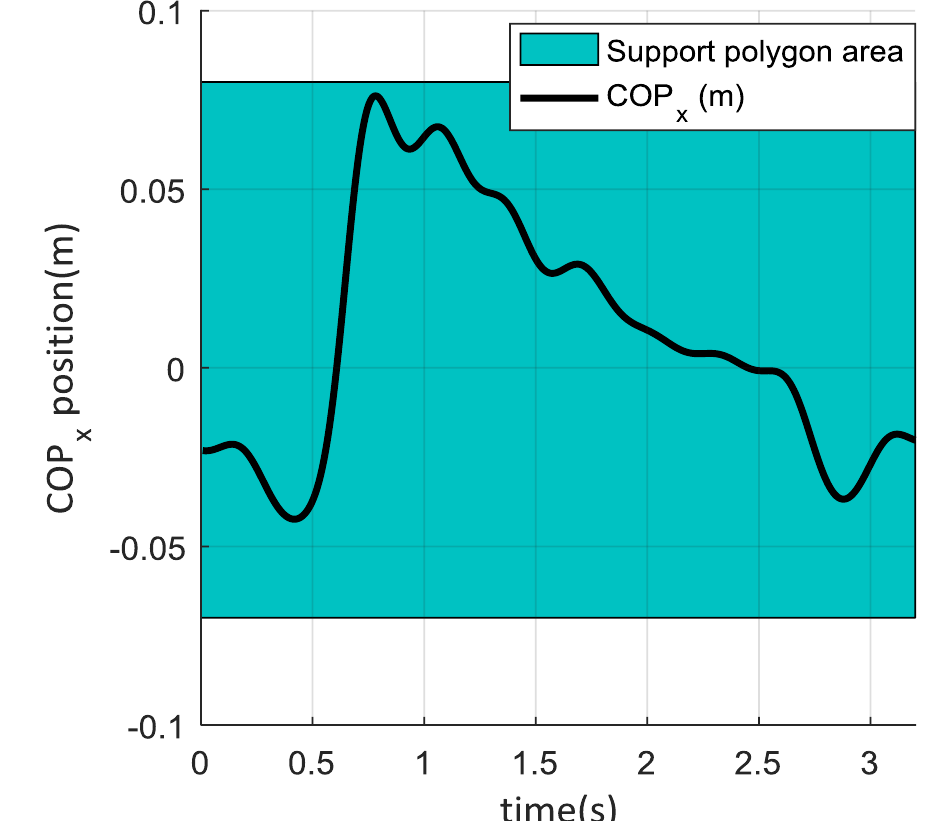}\par
\end{multicols}
\caption{Results of Push recovery of SURENA-Mini humanoid robot(Fig. \ref{fig4})}
 \label{fig6}
\end{figure*}

\subsection{Hip strategy}  
Hip strategy is equivalent to CMP balancing and is used for controlling the CP. In fact generally it can be handled by upper-body rotation(Hip joint) but it is not limited to this because if we assume that hip joint is locked and we don't have actuation in hip joint, we can generate angular momentum with knee or arm rotation. Therefore the hip strategy is not limited to use hip joint. Briefly it is equivalent to change angular momentum of the centroidal dynamic of a humanoid robot. By approximating the effect of centroidal angular momentum as a flywheel and by adding the torque of flywheel to the (\ref{eq9}), the (\ref{eq13}) will be modified as follows:

\begin{equation}
\begin{aligned}
\dt{\xi}_{x,error}=\omega_n({\xi_{x,error}}-\frac{{\tau}_{y,ankle}}{{mg}}-\frac{{\tau}_{y,hip}}{{mg}})\hspace{1.1cm}\\
\dt{\xi}_{y,error}=\omega_n({\xi_{y,error}}-\frac{{\tau}_{x,ankle}}{{mg}}+\frac{{\tau}_{x,hip}}{{mg}})\hspace{1.1cm}\\
\end{aligned}
  \label{eq15}
\end{equation} 
Considering the torque of hip joint as a control input (\ref{eq15}), the error of CP can be compensated  with the following PD controller:
  
\begin{equation}
\begin{aligned}
{\tau}_{y,hip}=k_p{\xi_{x,error}}+k_d{\dt{\xi}_{x,error}}\hspace{1.1cm}\\
{\tau}_{x,hip}=k_p{\xi_{y,error}}+k_d{\dt{\xi}_{y,error}}\hspace{1.1cm}\\
\end{aligned}
  \label{eq16}
\end{equation} 
It is noteworthy that input of the proposed controllers are torque of joints, however most of the humanoid robots are position-controlled. Therefore this controller is not implementable on  position-controlled humanoid robots.

\footnotetext{A video of the push recovery behavior experiment is available on https://youtu.be/2pPhKCSmwwo}

\section {PUSH  RECOVERY CONTROLLER}
In the third section a PD controller based on commanding the torque of ankle and hip joint is presented. Most of the conventional humanoid robots in practice are position-controlled and torque of each joint can be approximated for position controlled actuators with proportional control by directly setting the target angle of the actuator. This approximation is based on that every trajectory of torque can be generated with a trajectory of position. 

Therefor instead of the previous PD controller that is based on torque of joints, the feedback controller is modified to the following format: 
\begin{equation}
\begin{aligned}
{\theta}_{y,ankle}=k_p{\xi_{x,error}}+k_d{\dt{\xi}_{x,error}}\hspace{1.1cm}\\
{\theta}_{y,hip}=k_p{\xi_{x,error}}+k_d{\dt{\xi}_{x,error}}\hspace{1.1cm}\\
\end{aligned}
  \label{eq17}
\end{equation}
Note that this controller unlike the one is presented in (\ref{eq14}) is not model-based, however it works in reality very well.
In addition to the ankle and hip joints, we use the same control law to modulate arm position to generate additional effective angular momentum.
\begin{equation}
\begin{aligned}
{\theta}_{y,arm}=k_p{\xi_{x,error}}+k_d{\dt{\xi}_{x,error}}\hspace{1.1cm}\\
{\theta}_{y,elbow}=k_p{\xi_{x,error}}+k_d{\dt{\xi}_{x,error}}\hspace{1.1cm}\\
\end{aligned}
  \label{eq18}
\end{equation}

Therefore based on concept of hip and ankle strategy and inspired of the LIPM+flywheel, a feedback controller is presented that can be applicable on the position-controlled humanoid robots.

\begin{table}[t]
\caption{Characteristic of SURENA Mini humanoid robot}
\label{table_example}
\begin{center}
\begin{tabular}{@{\hskip 0.35in} c @{\hskip 0.5in} c @{\hskip 0.5in} c  @{\hskip 0.3in}}
\toprule
Variable & Value \\
\midrule
Height &53 cm\\
CoM Height &35 cm \\
 Total Mass & 3.6 \\
Foot Length & 15cm  \\
Foot Width& 8cm\\
Mass of each arm & 400g  \\
Total number of degrees of freedom &23  \\
\bottomrule
\end{tabular}
\end{center}
  \label{Tab1}
\end{table}
\subsection{Experiments on SURENA-Mini  Humanoid Robot}

In this section, we will show that the proposed algorithm can save the robot from falling under severe pushes. We do experiments on SURENA-Mini humanoid robot in situations that controller is active and is not active. SURENA-Mini humanoid robot is a kid-size 3D printed robot that is developed at Center of Advanced System and Technologies(CAST) at University of Tehran. SURENA-Mini is a position-controlled humanoid robot that does not have any force/torque sesnsors at its joints or feet and only has an IMU sensor on its  upper-body and an encoder at each joint. SURENA-Mini robot has position-controlled Dynamixel servos for actuators, which are controlled by an Intel embedded PC at a control frequency of 100Hz. The characteristics of SURENA-Mini humanoid robot is shown in  Table.\ref{Tab1}. Most of the conventional humanoid robots are position-controlled and have encoder at each joint and IMU at upper body. The encoders and IMU will be used for state estimation of CoM. Therefore the proposed algorithm is applicable to most of the available humanoid robots. 

The gain of PD controller is selected during experiments based on try and error to achieve the best performance of the push recovery controller on SURENA-Mini humanoid robot. 

Figures .\ref{fig4}, \ref{fig5} summarizes the result of applying the push on the robot in sagittal plane 
 with and without push recovery controller. As shown in the Fig. \ref{fig6}, when controller is not active the robot fails to recover the balance and falls, however when the controller is active the controller saves the robot from falling using ankle and hip strategy. For having uniform pushes on the robot the ball is released from the constant height  during the two experiments. 

 As shown in Fig. \ref{fig6}, the large push throws the CP out of the support polygon. The CoP remains in support polygon, however the CMP can go beyond the age of the foot and can push the CP to desired position. The angular momentum is generated by the rotation of hip and arm joints to move the CMP outside the support polygon for controlling the CP. Therefore, based on the experimental results, the regulation of angular momentum is so beneficial during push recovery.

\section{CONCLUSION AND FUTURE WORK}
In this paper, a push recovery controller based on CP concept using a feedback controller is developed. The core of the proposed feedback controller is based on a combined hip and ankle strategy to compensate the error of CP. The results showed that this controller is capable of rejecting severe pushes in the case where the stepping is not allowed. The main advantage of this controller is that it is  applicable on all of the conventional  position controlled humanoid robots with only an IMU sensor on the upper body and encoder at each joint and does  not need any force/torque sensors. The effectiveness of the proposed controller  was demonstrated by experimental implementation on SURENA-Mini humanoid robot. 

Despite all of the above advantages, there is a lot of room for improvement in this controller. First of all this controller does not present any priority in the strategies, however human uses the balancing strategies in an efficient way and prioritizes balancing strategies. Also, this controller does not consider the constraints of environment. Therefore adding a heuristic to consider the environment constraint is another suggestion for future works. Adding stepping strategy to this controller is now under development on SURENA-Mini humanoid robot.





\bibliographystyle{IEEEtran}
\bibliography{Master}

\end{document}